\documentclass[journal]{IEEEtran}

\ifCLASSINFOpdf
\else
   \usepackage[dvips]{graphicx}
\fi
\usepackage{url}
\usepackage{graphicx}
\usepackage{booktabs}
\usepackage{multirow}
\usepackage{makecell}
\usepackage[caption=false,font=normalsize,labelfont=sf,textfont=sf]{subfig}
\usepackage{algorithmic}
\usepackage{algorithm}
\usepackage{array}
\usepackage{amsmath}
\usepackage{amsfonts}
\usepackage{amssymb}
\usepackage{mathrsfs}
\usepackage{siunitx}
\usepackage{xcolor}

\begin{document}

\title{Dynamic Indoor Fingerprinting Localization based on Few-Shot Meta-Learning with CSI Images}

\author{ Jiyu Jiao, Xiaojun Wang, Chenpei Han, Yuhua Huang and Yizhuo Zhang 
\thanks{This work was supported in part by the National Key R$\&$D Program of China under Grant 2022YFC38010000. \textit{(Corresponding author: Xiaojun Wang)} }
\thanks{Jiyu Jiao, Chenpei Han, Yuhua Huang and Yizhuo Zhang are with the School of Information science and Engineering, Southeast University, Nanjing 214135, China (e-mail: jiyu$\_$jiao@seu.edu.cn;213201733@seu.edu.cn).}
\thanks{Xiaojun Wang is with National Mobile Communications Research Laboratory, School of Information science and Engineering, Southeast University, Nanjing 211100, China and Purple Mountain Laboratories, Nanjing 211111, China (e-mail: wxj@seu.edu.cn).}
}

\markboth{Journal of \LaTeX\ Class Files, Vol. , No. , xxx 202x}
{Shell \MakeLowercase{\textit{et al.}}: Bare Demo of IEEEtran.cls for IEEE Journals}
\maketitle
\vspace{-0.15cm}
\begin{abstract}
While fingerprinting localization is favored for its effectiveness, it is hindered by high data acquisition costs and the inaccuracy of static database-based estimates. Addressing these issues, this letter presents an innovative indoor localization method using a data-efficient meta-learning algorithm. This approach, grounded in the ``Learning to Learn'' paradigm of meta-learning, utilizes historical localization tasks to improve adaptability and learning efficiency in dynamic indoor environments. We introduce a task-weighted loss to enhance knowledge transfer within this framework. Our comprehensive experiments confirm the method's robustness and superiority over current benchmarks, achieving a notable 23.13\% average gain in Mean Euclidean Distance, particularly effective in scenarios with limited CSI data.
\end{abstract}
\vspace{-0.2cm}

\IEEEpeerreviewmaketitle
\vspace{-0.4cm}
\section{Introduction}
\IEEEPARstart{W}{ith} the rapid advancement of 5G and the Internet of Things (IoT), the demand for high-precision and low-latency positioning technologies is significantly increasing, playing a crucial role in driving the development of smart cities and shopping mall navigation etc\cite{lu2023tutorial}. Accurate positioning is not only key to precise data collection and efficient processing but also supports the stability and efficiency of wireless communication in aspects like resource allocation and pilot distribution. The continual enrichment of 5G/IoT low-power positioning terminals has propelled the commercial scaling of indoor positioning technologies. 

Traditional positioning methods like GPS have limited accuracy in indoor or dense urban environments, leading to extensive research into technologies like WiFi, Ultra-Wideband, and Bluetooth\cite{ruizhi2017indoor}. Specifically, WiFi positioning, with its widespread availability, lower cost, and moderate precision, demonstrates considerable potential in the field of indoor localization.

Geometric positioning methods\cite{chen2021carrier}\cite{menta2019performance} are accurate but costly and challenging for site selection and maintenance. In contrast, fingerprinting localization, not reliant on base station locations, noise statistics, or propagation parameters, approaches positioning as a classification problem, using signal characteristics like RSS\cite{guo2020hybrid} and CSI\cite{yang2013rssi} to build an offline training database. Traditional fingerprinting is split into deterministic methods, using similarity measures for position inference\cite{sensingmagicol}, and probabilistic methods, utilizing signal distributions for localization\cite{chen2020aoa}. The advent of machine learning algorithms, such as Support Vector Machines\cite{zhou2017device} and Deep Neural Networks\cite{comiter2018localization}, marks a new phase in deterministic approaches. However, these methods, though effective on static data, lack robustness to environmental changes, often resulting in inaccurate positioning from static databases.

Transfer learning (TL) aims to alleviate the scarcity of labeled data by borrowing knowledge from a different but related source domain, rich in labeled samples, to aid learning in a target domain with few (or no) labeled samples. Tsai et al.\cite{tsai2016learning} proposed a semi-supervised heterogeneous TL solution, selecting high-quality landmarks during the adaptation process to effectively reduce marginal and conditional distribution discrepancies between source and target domains. Besides, He et al.\cite{he2020heterogeneous} used a set of unlabeled correspondences between the source and target domains, enhancing the efficacy of TL.

Diverging from existing studies, this letter proposes a novel indoor localization scheme based on a data-efficient meta-learning algorithm, effectively addressing the robustness challenges in dynamic indoor environments and facilitating efficient cross-area localization, significantly enhancing data utilization and reducing collection costs. Our focus is on leveraging prior localization knowledge to strengthen the localization capability for target tasks. The key contributions of this letter include: 

\begin{enumerate}
	\item[1)]Proposing a few-shot meta-learning-based indoor localization approach with a versatile model designed for rapid adaptation to new environments and enhanced data efficiency, leveraging training knowledge from multiple datasets in various scenarios.
	\item[2)]For the substantial training tasks required by meta-learning algorithms, we propose a task-weighted loss to guide knowledge transfer.
	\item[3)]Extensive experiments validate the feasibility and robustness of our approach, demonstrating its superior performance over other benchmarks, achieving notable accuracy with a minimal number of samples.
\end{enumerate}
\vspace{-0.3cm}
\section{System Description}

\begin{figure*}[htbp]
	\centerline{\includegraphics[width=.85\textwidth]{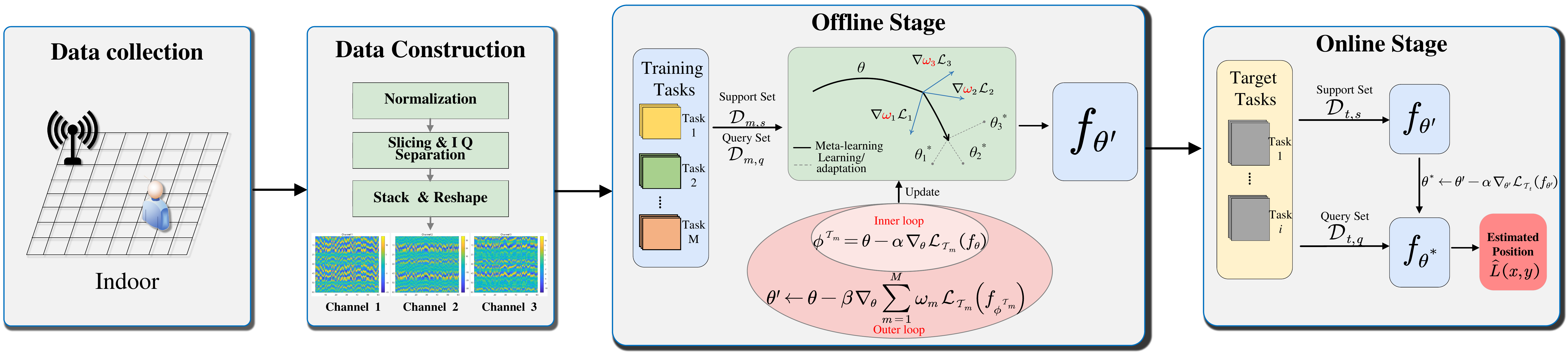}}
	\vspace{-0.15cm}
	\caption{Proposed few-shot meta-learning schematic for indoor localization.}
	\label{fig1}
\end{figure*}

\begin{figure}[h!]
	\centerline{\includegraphics[width=.48\textwidth]{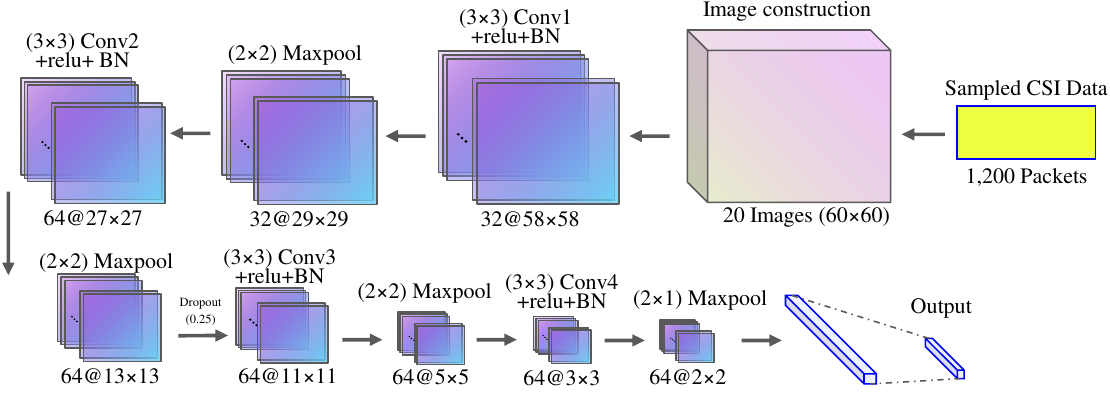}}
	\vspace{-0.15cm}
	\caption{Structure of inner model.}
	\label{fig2}
\end{figure}



\subsection{Problem Formulation}
We assume that indoor localization for scenarios occurring in subarea $ n_{sub} $ and posture $ f $ is defined as task $ \mathcal{T} $, which follows the task distribution $ p\left( \mathcal{T} \right) $. Let $\mathscr{F}=\left[ f_0,...,f_{i-1} \right]$ denote the set of postures, $ p\left( \mathcal{T} \right) =\sum_{n_{sub}=0}^{N_{sub}-1}{\sum_{f\in \mathscr{F}}{p\left( \mathcal{T}\mid n_{sub},f \right)}}\cdot p\left( n_{sub} \right) \cdot p\left( f \right) $, where $ p\left( \mathcal{T}\mid n_{sub},f \right) $denotes the probability distribution of the task under a specific area and posture, while $ p\left( n_{sub} \right) $ and $ p\left( f \right) $ represent the prior probabilities of the specific area and posture, respectively. The meta-task set is divided into support and query sets, thus the training task set can be defined as $\left\{ \mathcal{T}_m=\mathcal{D}_{m,s}\cup \mathcal{D}_{m,q} \right\} _{m=1}^{M}$, and the target task set as $\left\{ \mathcal{T}_t=\mathcal{D}_{t,s}\cup \mathcal{D}_{t,q} \right\} _{i=1}^{I}$. From task $\mathcal{T}$ sample $ K $ datapoints:
\begin{equation}
  \mathcal{D}_{s}=\left\{ \text{x}_{i}^{(k)},\text{y}_{i}^{(k)} \right\},i\in P_s,k\in \left[ 0,...,K-1 \right] 
	\label{eq33}
\end{equation}
from task $\mathcal{T}$ Sample $ K' $ datapoints:
\begin{equation}
  \mathcal{D}_{q}=\left\{ \text{x}_{j}^{(k)},\text{y}_{j}^{(k)} \right\},j\in P_q,k\in \left[ 0,...,K'-1 \right] 
\label{eq44}
\end{equation}
where $\text{x}$ represents the CSI fingerprint image and $ \text{y} $ denotes the corresponding two-dimensional coordinate label, forming the pair $ \left( csi,L \right) $. $\mathscr{P}_{n_{sub}}=P_s\cup P_q$ represent the set of reference points (RPs) in a subarea.

We view subarea localization as a regression task, where $\theta$ denotes the initial parameters of the model, and $f_\theta$ represents the trained model designed to learn the mapping from $csi$ to $L$. The meta-learning model's loss function, $\mathcal{L}_{\mathcal{T}}\left( f_{\theta}, \mathcal{D} \right) = \mathcal{L}^{task} = \mathcal{L}^{meta}$, assesses $f_{\theta}$'s learning performance, where $\mathcal{L}^{task}$ represents the task-level loss and $\mathcal{L}^{meta}$ the meta-level loss:
\begin{equation}
\mathcal{L}_{\mathcal{T}}\left( f_{\theta},\mathcal{D} \right) =\frac{1}{KN}\sum_{i=1}^{KN}{\left( L_i-\widehat{L}_i \right) ^2} 
\label{eq5}
\end{equation}
Where $ \mathcal{D} $ is the labeled dataset for a given task, $ KN $ is the total number of samples in either the support or query set, $ L_i $ and $ \widehat{L}_i $ are the actual and predicted coordinates, respectively.

As illustrated in Fig \ref{fig1}, in the offline phase, for all tasks, the objective is to minimize the total loss predicted over the query set $\mathcal{D}_{m,q}$ for the trained model $ f_{\phi ^{\mathcal{T}_m}} $ after training on the support set $\mathcal{D}_{m,s}$ according to Equation \ref{eq7} and \ref{eq8}. This essentially involves finding the optimal initial model parameters $\theta '$ to enable rapid adaptation in another task. In the online phase, the new initial parameters $\theta '$ and a small amount of the support set $ \mathcal{D}_{t,s} $ of the target task are used for a few steps of rapid inference learning.
\begin{equation}
		\phi ^{\mathcal{T}_m}=\theta -\alpha \nabla _{\theta}\mathcal{L}_{\mathcal{T}_m}\left( f_{\theta} \right)
	\label{eq7}
\end{equation}
\begin{equation}
		\theta ' \gets \ \theta -\beta \nabla _{\theta}\sum_{m=1}^M{\omega _m\mathcal{L}_{\mathcal{T}_m}\left( f_{\phi ^{\mathcal{T}_m}} \right)}
	\label{eq8}
\end{equation}

We designed a versatile network model as depicted in Fig. \ref{fig2}, comprising four convolutional layers and one fully connected layer. The input is a CSI image of size $ N_t\times 2\mathscr{N}$, where $ \mathscr{N} $ represents the number of subcarriers, $ N_t $ represents the number of continuously collected CSI packets. The output is the estimated two-dimensional coordinates. The localization error is quantified using the Mean Euclidean Distance (MED):
\begin{equation}
	MED=\frac{1}{KN_q}\sum_{i=1}^{KN_q}{\sqrt{\sum_{j=1}^d{\left( f_{\theta}\left( csi_i \right) _j-L_{ij} \right) ^2}}} 
	\label{eq66}
\end{equation}
Where $KN_q$ is the total number of test samples, $d$ is the dimensionality of the coordinates, and $ f_{\theta}\left( csi_i \right)_j $ is the predicted value in the $ j $th dimension for the $ i $th sample.
\vspace{-0.45cm}
\subsection{Enhancing Meta-Learning with Task Importance Weighting via Wasserstein Distance (W-Dis)}
In traditional meta-learning frameworks, different tasks are usually considered equally contributive to learning new tasks. However, we argue that tasks with distributions closer to the target task should be more influential. Utilizing statistical measures like W-Dis \cite{rubner2000earth} or KL divergence to assess distribution differences, we propose an importance weighting method that employs W-Dis loss to gauge each task's relevance in learning new tasks. This method prioritizes tasks similar to the target, facilitating enhanced knowledge transfer and quicker adaptation in meta-learning.

W-Dis quantifies the minimum average distance required to move data from distribution $ p(x) $ to $ q(y) $. When distributions $ p $ and $ q $ are far apart with no overlap, the KL divergence becomes meaningless and the JS divergence converges to a constant. This is problematic in learning algorithms as it leads to gradient vanishing. W-Dis addresses this issue. Its formalized expression is as follows:
\begin{equation}
	W\left( p,q \right) =\underset{\gamma \thicksim \prod{\left( p,q \right)}}{\text{inf}}\text{E}_{x,y\thicksim \gamma}\left[ \lVert x-y \rVert \right]
\label{eq66}
\end{equation}	
where $\prod{\left( p,q \right)} $ represents the set of all possible joint distributions combining distributions $p$ and $q$. For each possible joint distribution $\gamma$, one can sample pairs $ \left(x,y\right)\thicksim \gamma $ to obtain samples $ x $ and $ y $, and compute the distance $\lVert x-y \rVert $ between these samples. Therefore, the expected value of the sample pair distances under the joint distribution $ \gamma $ can be calculated as $ \text{E}_{x,y\thicksim \gamma}\left[ \lVert x-y \rVert \right] $. The lower bound of this expected value across all possible joint distributions constitutes the W-Dis.

Using the aforementioned method, calculate the W-Dis $W_{tm}$ between the target task and training tasks. Then, determine the corresponding contribution weights $\omega ' _{tm}$ using the softmax function:
\begin{equation}
	\mathcal{W}=\left[ \omega '_{t1},...,\omega ' _{tM} \right] =softmax\left( \left[ -W_{t1},...,-W_{tM} \right] ^{\top} \right)  
	\label{eq77}
\end{equation}

Considering that values post-softmax normalization are confined between 0 and 1, resulting in notably small weights $\omega ' _{tm}$ and consequently small feedback loss values $\sum{\omega \mathcal{L}_{\mathcal{T}}\left( f_{\phi ^{\mathcal{T}}} \right)}$, the following method is employed to derive the final weights:
\begin{equation}
	\omega =\frac{\omega '_{tb}}{\sum_{b=1}^B{\omega '_{tb}}}  
	\label{eq77}
\end{equation}
Where $ B $ represents the batch size during training.

\begin{algorithm}[htbp]
	\caption{Meta-Learning Indoor Localization on Few-Shot Image Samples.}\label{alg:alg1}
	\begin{algorithmic}
		\REQUIRE Task set $ \left\{ \mathcal{T}_m \right\} _{m=1}^{M} $and a target task $ \mathcal{T}_t $, a meta-model $ f $ and initial parameters $ \theta $. datapoints number $ K $, learning rates $ \alpha $ and $ \beta $, iteration count $ epochs $, meta batch size $ B $, task-level inner update steps $N_{in} $, and finetuning update steps $N_{fi} $.		
		\ENSURE The estimated location for the target task $ \widehat{L}_i $ .
		\STATE \textbf{Offline stage:} 
		\STATE Randomly initialize $ \theta $;
		\WHILE {not done}
		\STATE Sample batch of tasks $ \left\{ \mathcal{T}_m \right\} _{m=1}^{M} \sim p\left( \mathcal{T} \right) $;
		\FOR{\textbf{all} $ \mathcal{T}_m $} 
		\STATE Sample $ K $ datapoints $ \mathcal{D}_{m,s}=\left\{ \text{x}_{i}^{(k)},\text{y}_{i}^{(k)} \right\}  $ from $  \mathcal{T}_m $;
		\STATE Evaluate $ \nabla _{\theta}\mathcal{L}_{\mathcal{T}_m}\left( f_{\theta} \right) $ using $ \mathcal{D}_{m,s} $ and $ \mathcal{L}_{\mathcal{T}_m} $;
		\STATE Compute adapted parameters with gradient descent:
		\STATE $ \phi ^{\mathcal{T}_m}=\theta -\alpha \nabla _{\theta}\mathcal{L}_{\mathcal{T}_m}\left( f_{\theta} \right)  $;
		\STATE Sample $ K' $ datapoints $ \mathcal{D}_{m,q}=\left\{ \text{x}_{j}^{(k)},\text{y}_{j}^{(k)} \right\}  $ from $  \mathcal{T}_m  $ for the meta-update;
		\ENDFOR
		\STATE Evaluating the importance vector $ \mathcal{W}=\left[ \omega _1,...,\omega _M \right]  $ of training tasks using the target task based on eq( );
		\STATE Update $ \theta ' \gets \theta -\beta \nabla _{\theta}\sum_{m=1}^M{\omega _m\mathcal{L}_{\mathcal{T}_m}\left( f_{\phi ^{\mathcal{T}_m}} \right)}  $ using each $ \mathcal{D}_{m,q} $ and $ \mathcal{L}_{\mathcal{T}_m} $;
		\ENDWHILE
		\STATE \textbf{Online stage:} 
		\WHILE {not done}
		\STATE Use the support set of the test task $ \mathcal{D}_{t,s} $ to finetune the model $ f $:
		\STATE $ \theta ^*\gets \theta '-\alpha \nabla _{\theta '}\mathcal{L}_{\mathcal{T}_t}\left( f_{\theta '} \right)  $;
		\STATE Make predictions using the query set of the test task $ \mathcal{D}_{t,q} $:
		\STATE $ \widehat{L}=f_{\theta ^*}\left( \mathbf{x}_{i}^{\left( k \right)} \right) ,\mathbf{x}_{i}^{\left( k \right)}\in \mathcal{D}_{t,q} $.
		\ENDWHILE
	\end{algorithmic}
	\label{alg1}
\end{algorithm}
\vspace{-0.1cm}
\subsection{Image construction}
Let $ \mathbb{C}\mathbb{S}\mathbb{I}_i=\left( \mathcal{H}_{i1},...,\mathcal{H}_{ij},...,\mathcal{H}_{iJ} \right)  $ denote the CSI fingerprints of the $ i $th RP, where $ \mathcal{H}_{ij} $ is the CSI from the $j$th data packet.

For a  Single Input Multiple Output(SIMO) indoor localization system with a single-antenna transmitter and a three-antenna receiver, the frequency response of a single packet sample is given as:
\begin{equation}
		\mathcal{H}=\left\{ h_{i}^{\left( n \right)}|i\in \left[ 1, \mathcal{N}_r  \right] ,n\in \left[ 1,\mathscr{N} \right] \right\} 
	\label{eq13}
\end{equation}
\begin{equation}
		h_{i}^{\left( n \right)}=I_{i}^{\left( n \right)} + jQ_{i}^{\left( n \right)} =\left| h_{i}^{\left( n \right)} \right|e^{j\angle h_{i}^{\left( n \right)}},\ n\in \left[ 1,\mathscr{N} \right]
	\label{eq14}
\end{equation}
where $ \mathcal{N}_r $ is the number of receiving antennas, $ \left| h_{i}^{\left( n \right)} \right| $ and $ \angle h_{i}^{\left( n \right)} $ are the amplitude and phase of the $ n $th subcarrier, respectively, and $ I_{i}^{\left( n \right)} $ and $ Q_{i}^{\left( n \right)} $ are the in-phase and quadrature components, respectively.

The in-phase and quadrature components of a single sample's frequency response are separated and reshaped along the y-axis to form $\boldsymbol{h}_i$: 
\begin{equation}
		\boldsymbol{h}_i=\left[ \begin{array}{c}
			\boldsymbol{I}_i\\
			\boldsymbol{Q}_i\\
		\end{array} \right] =\left[ I_{i}^{\left( 1 \right)},...,I_{i}^{\left( \mathscr{N} \right)},Q_{i}^{\left( 1 \right)},...,Q_{i}^{\left( \mathscr{N} \right)} \right] ^{\top}
	\label{eq10}
\end{equation}
the $ \boldsymbol{h}_{i}^{n_t} $ from $ N_t $ packets on the same antenna are then concatenated along the x-axis to form $ \boldsymbol{csi}^{*} $. Finally, $ \boldsymbol{csi}^{*} $ from different antennas are treated as separate channels and reshaped into the $ csi $ data:
\begin{equation}
		\boldsymbol{csi}^{*}=\left[ \boldsymbol{h}_{i}^1,...,\boldsymbol{h}_{i}^{n_t},...,\boldsymbol{h}_{i}^{N_t} \right]
	\label{eq16}
\end{equation}
\begin{equation}
		csi=\boldsymbol{csi}^R\bigoplus{\boldsymbol{csi}^G\bigoplus{\boldsymbol{csi}^B}}
	\label{eq17}
\end{equation}
\begin{figure}
	\centerline{\includegraphics[width=.3\textwidth]{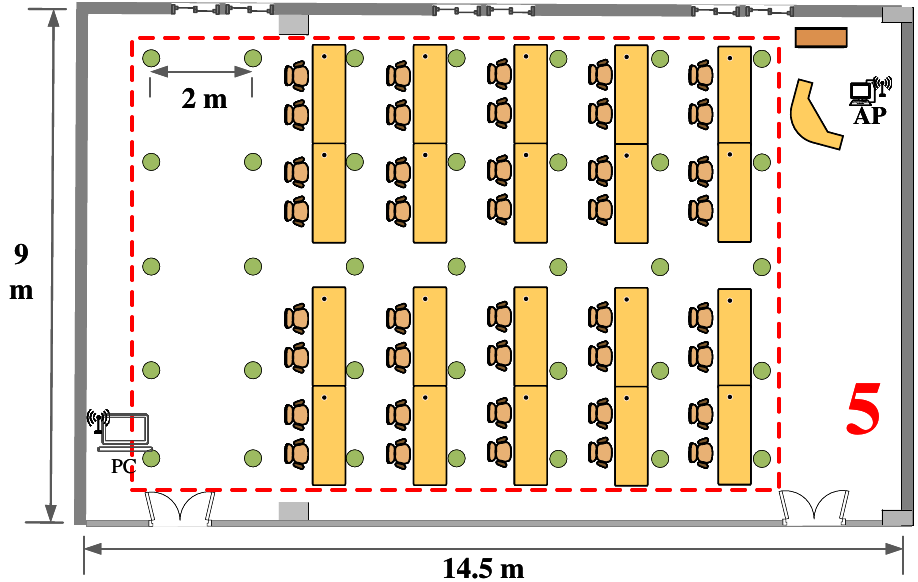}}
	\vspace{-0.15cm}
	\caption{Planar Layout of the Experimental Scenario B(RPs denoted by green dots).}
	\label{fig3}
\end{figure}
\vspace{-1cm}
\section{Experiment Results} 
\subsection{Experimental Setup}

\emph{1) Experimental scenario A\cite{wei2023meta}:} We utilized the experimental setup described in\cite{wei2023meta}, which includes three distinct areas - a corridor, laboratory, and hall, and we divided them into areas 1 through 4. CSI data was collected in each area using a uniform setup of a commercial wireless AP as the transmitter and a laptop with Intel 5300 NIC as the receiver. Five distinct human activities were recorded at each RP, leading to 20 indoor localization tasks, with RPs spaced 0.6 $m$ apart.

\emph{2) Experimental scenario B:} Fig. \ref{fig3} depicts a second experimental scene in area 5, a classroom in Nanjing's China Wireless Valley office building. Here, five human postures (standing, walking, squatting, multiple people standing) were recorded at each RP, creating five tasks, with RPs 2 meters apart. Both the Lenovo desktop AP and Lenovo laptop receiver, equipped with Intel 5300 NICs and Linux 802.11n CSI Tool, collected 2000 CSI packets over two seconds per RP. The setup operated in Monitor mode on channel 116, with the AP transmitting at a rate of 1000 packets/s using LORCON version 1 for packet injection.

\begin{figure}[htbp]
	\centerline{\includegraphics[width=0.62\columnwidth]{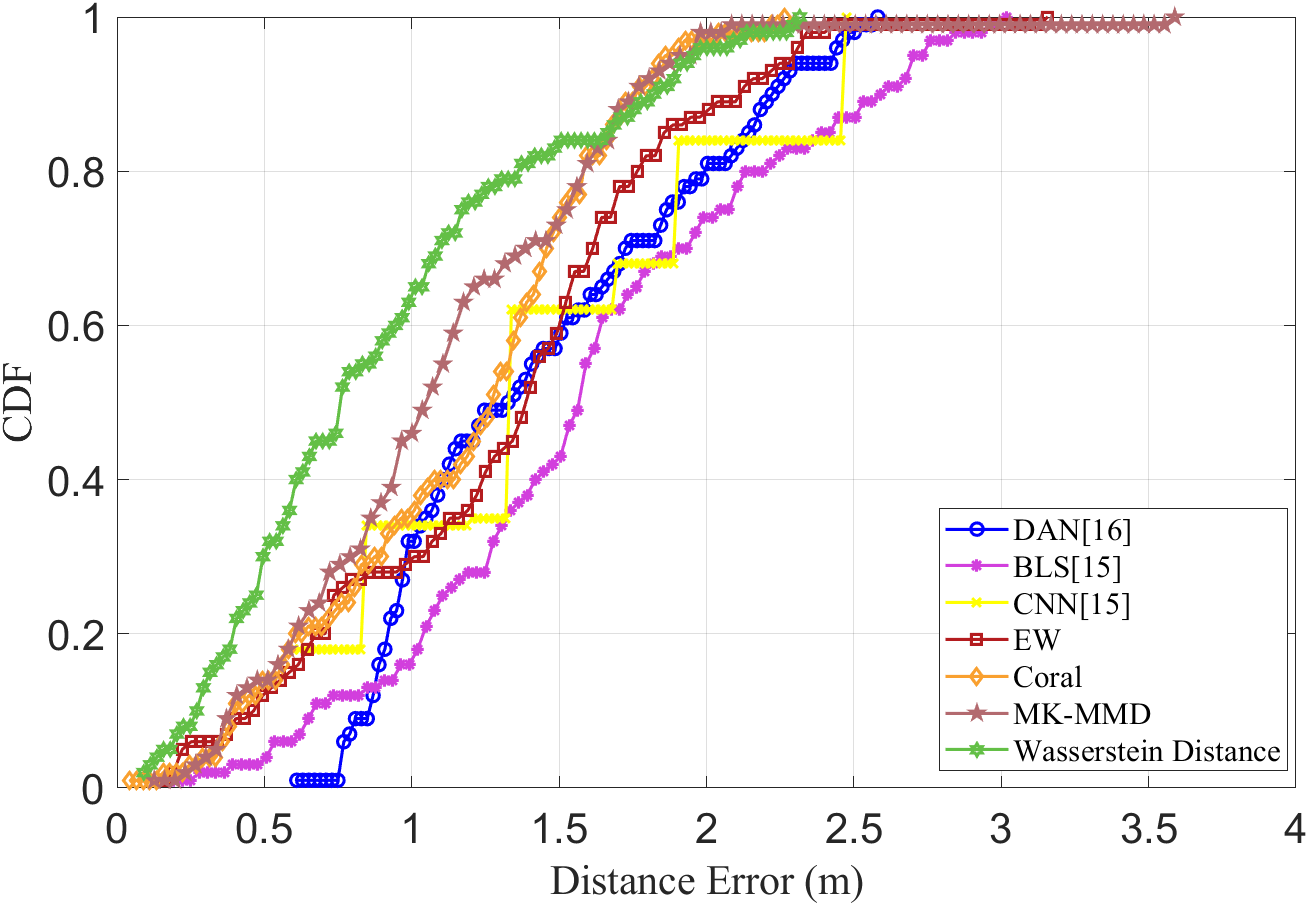}}
	\vspace{-0.22cm}
	\caption{CDF of Localization Errors for Different Algorithms (area 3).}
	\label{fig4}
\end{figure}
\vspace{-0.05cm}
\setlength{\tabcolsep}{2pt}
\renewcommand{\arraystretch}{0.2}
\begin{table}[htbp]
	\centering
	\caption{The comparison of MED across all areas using the experience from other tasks between the proposed method and existing methods.}
	\label{tab1}
	\begin{tabular}{cccccccc} 
		\toprule
		\multirow{2}{*}{Area} & \multicolumn{3}{c}{Existing Methods} & \multicolumn{4}{c}{Proposed Meta Learing Metheds} \\ \cline{2-8}
		
		& \makecell[c]{CNN\cite{zhu2022intelligent}}  & \makecell[c]{BLS\cite{zhu2022intelligent}}  & \makecell[c]{DAN\cite{xiang2021self}}   & \makecell[c]{W-Dis} & \makecell[c]{Coral Loss}& \makecell[c]{MK-MMD}& \makecell[c]{NW} \\ \midrule
		
		1& 1.307   & 1.500  & 1.565    &  1.241  & 1.151  & 1.229  & 1.303    \\ 
		2 & 2.848   & 3.032  & 2.164    &  1.560  & 2.066  & 1.647  & 2.264   \\ 
		3 & 1.419   & 1.607  & 1.389    &  0.901  & 1.434  & 1.108  & 1.319 \\ 
		4 & 1.555  & 1.566 & 1.769  &  1.358  & 1.392  & 1.589  & 1.412         \\ 
		5 & 5.596   & 5.903 & 5.012   &  3.678  & 4.220   & 4.426  & 3.934\\
		Mean & 2.545  & 2.722  & 2.380   & 1.747  &2.052  & 1.999 & 2.046  \\ \bottomrule
	\end{tabular}
\end{table}
\begin{figure}[htbp]
	\centerline{\includegraphics[width=0.58\columnwidth]{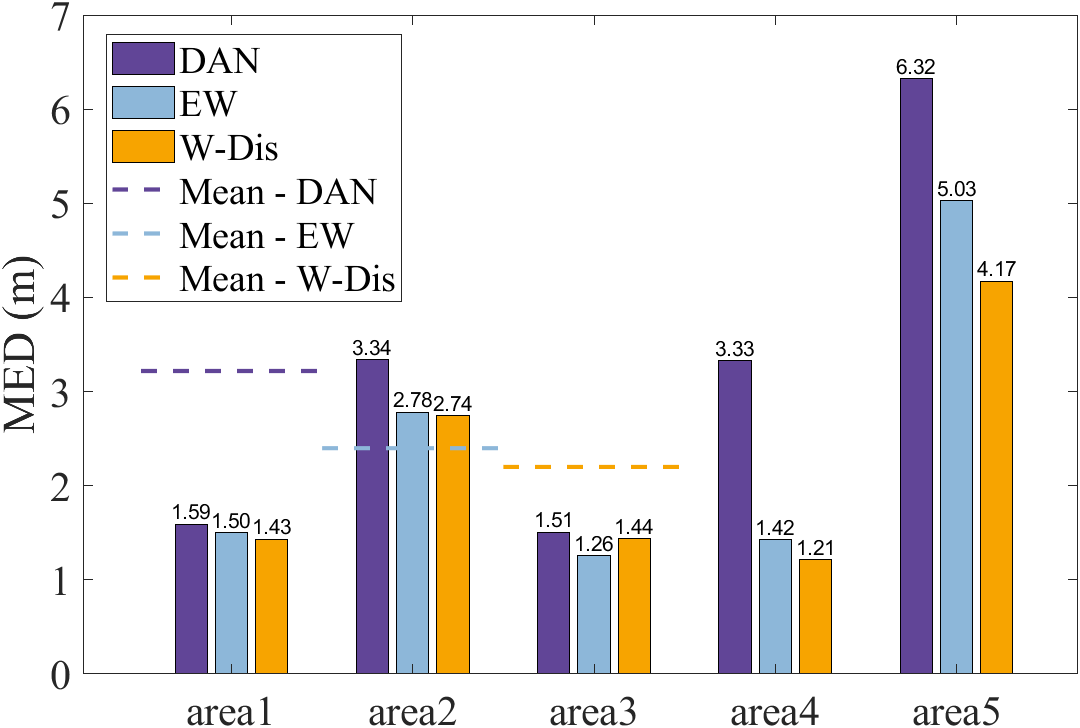}}
	\vspace{-0.17cm}
	\caption{Comparison of MED Across Different Algorithms (1-Input 1-Output).}
	\label{fig5}
\end{figure}

Our method, developed in a PyTorch framework and accelerated with an NVIDIA Tesla V100 GPU, involved setting up $ N_{sub}=5 $, each with $ p_{ap}=1 $ AP, and $ N_t=60 $ packets. We used $ \mathscr{N}=30 $ subcarriers. The support set RPs, $P_s$, comprised $ 80\% $ of each area's RPs, and the query set, $ P_q $, the remaining $ 20\% $. For both training and prediction, we set the number of datapoints for the support and query sets to $ K=K '=20 $. The learning rates were $ \alpha =0.01 $ and $\beta =0.001 $, with $epochs=6$, $B=4 $, $N_{in}=5 $, and $N_{fi}=13 $.    
\vspace{-0.3cm}
\subsection{Performance Analysis}
To evaluate our method, we conducted comparisons with the CNN\cite{zhu2022intelligent}, BLS\cite{zhu2022intelligent}, and a TL method using DAN\cite{xiang2021self}, referring to our basic meta-learning approach as NW (Non-Weighting). Among 25 tasks, one was randomly chosen as the target, with the others for training. As shown in Table \ref{tab1}, our W-Dis-based method achieved a 1.747m MED, significantly outperforming existing methods. On average, our task-weighted meta-learning method surpassed others by $23.13\%$ in localization error accuracy, with W-Dis-based showing the best performance, exceeding other proposed methods by $14.04\%$. Notably, it excelled in area 5, underlining its adaptability. Fig. \ref{fig4} demonstrates our method's superiority, with a CDF of $63.5\%$ at 1 m localization error.


Table \ref{tab1tab2} shows area 5's localization performance using various training tasks. Training across two areas, four areas and all areas achieved MEDs of 5.280m, 4.060m, and 4.650m, respectively. This demonstrates our method's ability to enhance target task localization by leveraging experience from other tasks while reducing the impact of less relevant tasks. As inferred from Fig. \ref{fig5} and Table \ref{tab1}, SISO systems exhibit lesser localization performance compared to SIMO systems, attributed to the latter's richer channel information.

\vspace{-0.05cm}
\setlength{\tabcolsep}{25pt}
\renewcommand{\arraystretch}{0.2}
\begin{table}[htbp]
	\centering
	\caption{Localization Performance of Area 5 Using Different Training Tasks}
	\label{tab1tab2}
	\begin{tabular}{ccc} 
		\toprule
		\makecell[c]{Area} & NW & W-Dis \\
		\midrule
		
		1+2 & 5.137   & 5.280     \\ 
		2+3+4 & 4.855   & 4.906   \\ 
		1+2+3+4 & 5.082   & 4.820  \\ 
		1+2+3+5 & 4.560  & 4.060       \\ 
		1+2+3+4+5 & 5.350   & 4.650  \\
		Mean & 4.996  & 4.743   \\ \bottomrule
	\end{tabular}
\end{table}
\begin{figure}[htbp]
	\centerline{\includegraphics[width=0.58\columnwidth]{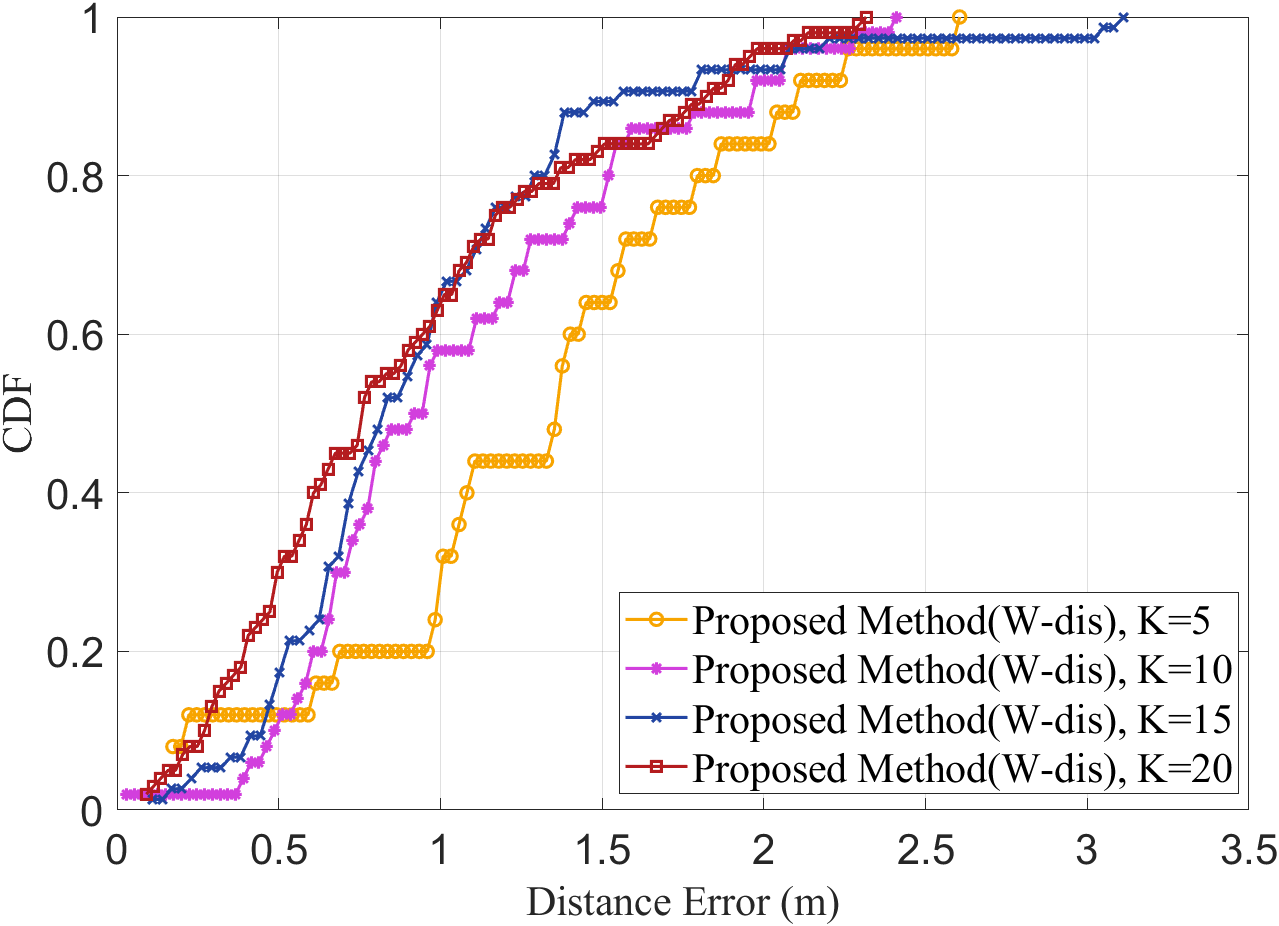}}
	\vspace{-0.22cm}
	\caption{CDF of Localization Error Under Different K-Values.}
	\label{fig6}
\end{figure}
\begin{figure}[htbp]
	\centerline{\includegraphics[width=0.58\columnwidth]{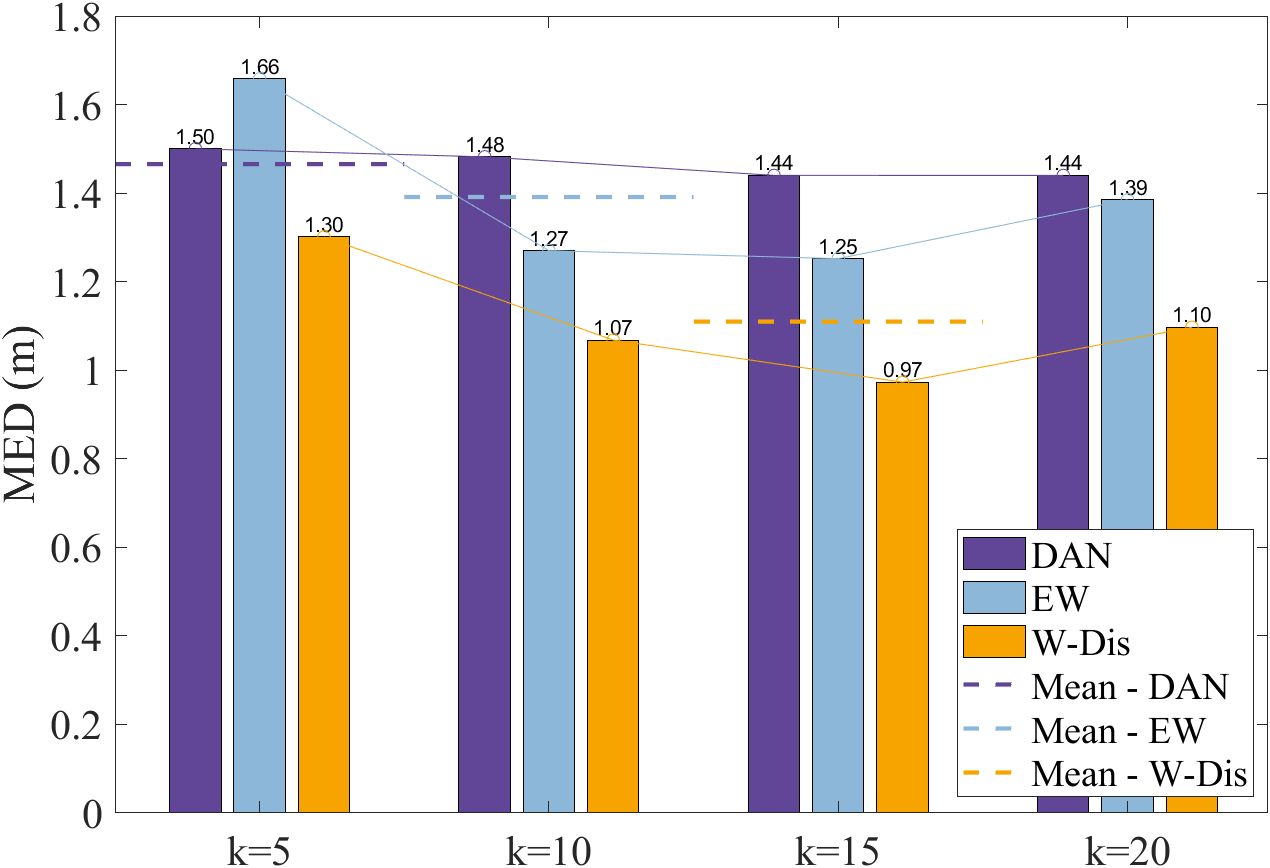}}
	\vspace{-0.17cm}
	\caption{The Impact of $ K $-Value on Localization Performance.}
	\label{fig7}
\end{figure}

Figures \ref{fig6} and \ref{fig7} show how varying the number of training samples, $K$, affects localization performance. Both meta-learning and TL methods improve as $K$ increases, peaking at $K=15$. Our method consistently outperforms others, demonstrating that effective new task localization can be achieved with experiential knowledge transfer, even with a limited number of samples.
\vspace{-0.2cm}
\section{Concluding Remarks}
This letter introduces a novel few-shot meta-learning-based indoor fingerprinting localization method, allowing a generic model to swiftly adapt to new tasks using knowledge from prior tasks. We utilize a task-weighting mechanism for effective knowledge transfer, enhancing learning for the target task. Our extensive experiments confirm our method's effectiveness and robustness, outperforming existing approaches and achieving efficient localization with only minimal samples, marking a significant advance in indoor localization.



\bibliographystyle{IEEEtran}
\bibliography{SPLref}

\begin{thebibliography}{10}
\providecommand{\url}[1]{#1}
\csname url@samestyle\endcsname
\providecommand{\newblock}{\relax}
\providecommand{\bibinfo}[2]{#2}
\providecommand{\BIBentrySTDinterwordspacing}{\spaceskip=0pt\relax}
\providecommand{\BIBentryALTinterwordstretchfactor}{4}
\providecommand{\BIBentryALTinterwordspacing}{\spaceskip=\fontdimen2\font plus
\BIBentryALTinterwordstretchfactor\fontdimen3\font minus
  \fontdimen4\font\relax}
\providecommand{\BIBforeignlanguage}[2]{{%
\expandafter\ifx\csname l@#1\endcsname\relax
\typeout{** WARNING: IEEEtran.bst: No hyphenation pattern has been}%
\typeout{** loaded for the language `#1'. Using the pattern for}%
\typeout{** the default language instead.}%
\else
\language=\csname l@#1\endcsname
\fi
#2}}
\providecommand{\BIBdecl}{\relax}
\BIBdecl

\bibitem{lu2023tutorial}
H.~Lu, Y.~Zeng, C.~You, Y.~Han, J.~Zhang, Z.~Wang, Z.~Dong, S.~Jin, C.-X. Wang,
  T.~Jiang \emph{et~al.}, ``A tutorial on near-field xl-mimo communications
  towards 6g,'' \emph{arXiv preprint arXiv:2310.11044}, 2023.

\bibitem{ruizhi2017indoor}
C.~Ruizhi and C.~Liang, ``Indoor positioning with smartphones: The
  state-of-the-art and the challenges,'' \emph{Acta Geodaetica et Cartographica
  Sinica}, vol.~46, no.~10, p. 1316, 2017.

\bibitem{chen2021carrier}
L.~Chen, X.~Zhou, F.~Chen, L.-L. Yang, and R.~Chen, ``Carrier phase ranging for
  indoor positioning with 5g nr signals,'' \emph{IEEE Internet of Things
  Journal}, vol.~9, no.~13, pp. 10\,908--10\,919, 2021.

\bibitem{menta2019performance}
E.~Y. Menta, N.~Malm, R.~J{\"a}ntti, K.~Ruttik, M.~Costa, and K.~Lepp{\"a}nen,
  ``On the performance of aoa--based localization in 5g ultra--dense
  networks,'' \emph{Ieee Access}, vol.~7, pp. 33\,870--33\,880, 2019.

\bibitem{guo2020hybrid}
X.~Guo, N.~Ansari, L.~Li, and L.~Duan, ``A hybrid positioning system for
  location-based services: Design and implementation,'' \emph{IEEE
  Communications Magazine}, vol.~58, no.~5, pp. 90--96, 2020.

\bibitem{yang2013rssi}
Z.~Yang, Z.~Zhou, and Y.~Liu, ``From rssi to csi: Indoor localization via
  channel response,'' \emph{ACM Computing Surveys (CSUR)}, vol.~46, no.~2, pp.
  1--32, 2013.

\bibitem{sensingmagicol}
O.~W. Sensing, ``Magicol: Indoor localization using pervasive magnetic field
  and opportunistic wifi sensing.''

\bibitem{chen2020aoa}
L.~Chen, I.~Ahriz, and D.~Le~Ruyet, ``Aoa-aware probabilistic indoor location
  fingerprinting using channel state information,'' \emph{IEEE internet of
  things journal}, vol.~7, no.~11, pp. 10\,868--10\,883, 2020.

\bibitem{zhou2017device}
R.~Zhou, X.~Lu, P.~Zhao, and J.~Chen, ``Device-free presence detection and
  localization with svm and csi fingerprinting,'' \emph{IEEE Sensors Journal},
  vol.~17, no.~23, pp. 7990--7999, 2017.

\bibitem{comiter2018localization}
M.~Comiter and H.~Kung, ``Localization convolutional neural networks using
  angle of arrival images,'' in \emph{2018 IEEE global communications
  conference (GLOBECOM)}.\hskip 1em plus 0.5em minus 0.4em\relax IEEE, 2018,
  pp. 1--7.

\bibitem{tsai2016learning}
Y.-H.~H. Tsai, Y.-R. Yeh, and Y.-C.~F. Wang, ``Learning cross-domain landmarks
  for heterogeneous domain adaptation,'' in \emph{Proceedings of the IEEE
  conference on computer vision and pattern recognition}, 2016, pp. 5081--5090.

\bibitem{he2020heterogeneous}
Y.~He, X.~Jin, G.~Ding, Y.~Guo, J.~Han, J.~Zhang, and S.~Zhao, ``Heterogeneous
  transfer learning with weighted instance-correspondence data,'' in
  \emph{Proceedings of the AAAI Conference on Artificial Intelligence},
  vol.~34, no.~04, 2020, pp. 4099--4106.

\bibitem{rubner2000earth}
Y.~Rubner, C.~Tomasi, and L.~J. Guibas, ``The earth mover's distance as a
  metric for image retrieval,'' \emph{International journal of computer
  vision}, vol.~40, pp. 99--121, 2000.

\bibitem{wei2023meta}
W.~Wei, J.~Yan, X.~Wu, C.~Wang, and G.~Zhang, ``A meta-learning approach for
  device-free indoor localization,'' \emph{IEEE Communications Letters},
  vol.~27, no.~3, pp. 846--850, 2023.

\bibitem{zhu2022intelligent}
X.~Zhu, W.~Qu, X.~Zhou, L.~Zhao, Z.~Ning, and T.~Qiu, ``Intelligent
  fingerprint-based localization scheme using csi images for internet of
  things,'' \emph{IEEE Transactions on Network Science and Engineering},
  vol.~9, no.~4, pp. 2378--2391, 2022.

\bibitem{xiang2021self}
C.~Xiang, S.~Zhang, S.~Xu, and G.~C. Alexandropoulos, ``Self-calibrating indoor
  localization with crowdsourcing fingerprints and transfer learning,'' in
  \emph{ICC 2021-IEEE International Conference on Communications}.\hskip 1em
  plus 0.5em minus 0.4em\relax IEEE, 2021, pp. 1--6.

\end{thebibliography}

%
%
%

\end{document}